\documentclass{article}

   \PassOptionsToPackage{numbers, compress}{natbib}

\usepackage[preprint]{neurips_2020}

\usepackage[utf8]{inputenc} 
\usepackage[T1]{fontenc}    
\usepackage{hyperref}       
\usepackage{url}            
\usepackage{booktabs}       
\usepackage{amsfonts}       
\usepackage{nicefrac}       
\usepackage{microtype}      
\usepackage{bm}
\usepackage{amsmath}
\usepackage{comment}
\newcommand{\norm}[1]{\left\lVert#1\right\rVert}
\usepackage{float}
\usepackage{graphicx}
\usepackage{float}
\usepackage{multirow}
\usepackage{array, booktabs, makecell}
\usepackage{arydshln}
\usepackage{caption}

\usepackage{mathtools}

\DeclarePairedDelimiter\floor{\lfloor}{\rfloor}

\title{Anomaly Detection with Tensor Networks}
 \author{%
  Jinhui Wang\thanks{Work done while a resident of X.} \\
  Stanford University\\
  Stanford, CA 94305, USA \\
  \texttt{wangjh97@stanford.edu} \\
  \And 
  Chase Roberts \\
  X - The Moonshot Factory\\
  Mountain View, CA 94043, USA \\
  \texttt{chaseriley@google.com} \\
 \And 
  Guifre Vidal \\
  X - The Moonshot Factory\\
  Mountain View, CA 94043, USA \\
  \texttt{guifre@google.com} \\
  \And 
  Stefan Leichenauer  \\
  X - The Moonshot Factory\\
  Mountain View, CA 94043, USA \\
  \texttt{sleichenauer@google.com} \\
}
\begin{document}

\graphicspath{ {./figures/} }

\maketitle

\begin{abstract}
Originating from condensed matter physics, tensor networks are compact representations of high-dimensional tensors. In this paper, the prowess of tensor networks is demonstrated on the particular task of one-class anomaly detection. We exploit the memory and computational efficiency of tensor networks to learn a linear transformation over a space with dimension exponential in the number of original features. The linearity of our model enables us to ensure a tight fit around training instances by penalizing the model's global tendency to a predict normality via its Frobenius norm---a task that is infeasible for most deep learning models. Our method outperforms deep and classical algorithms on tabular datasets and produces competitive results on image datasets, despite not exploiting the locality of images.
 
\end{abstract}

\section{Introduction}
Anomaly detection (AD) entails determining whether a data point comes from the same distribution as a prior set of normal data. Anomaly detection systems are used to discover credit card fraud, detect cyber intrusion attacks and identify cancer cells. Since normal examples are readily available while anomalies tend to be rare in production environments, we consider the semi-supervised or one-class setting where only normal instances are present in the training set. It is important to remark that the outlier space is often much larger than the inlier space, though anomalous observations are uncommon. For instance, the space of normal dog images is sparse in the space of anomalous non-dog images. This discrepancy between data availability and space sizes makes anomaly detection hard, as a model must account for the entire input space despite only having information about a minuscule subspace. Deep learning models generally struggle with this challenge since it is impractical to manage their behavior over the entire input space. Linear models, however, do not face such a difficulty.

To gain control over our model's behavior on the entire input space, we employ a linear transformation as its main component and subsequently penalize its Frobenius norm. However, this transformation has to be performed over an exponentially large feature space for our model to be expressive---an impossible task with full matrices. Thus, we leverage tensor networks as sparse representations of such large matrices. All-in-all, our model is an end-to-end anomaly detector for general data that produces a normality score via its decision function $\mathcal{D}$. Our novel method is showcased on several tabular and image datasets. We attain significant improvements over prior methods on tabular datasets and competitive results on image datasets, despite not exploiting the locality of image pixels.

\section{Related Work}
\citet{miles} first demonstrated the potential of tensor networks in classification tasks, using the well-known density matrix renormalization group algorithm from computational physics \citep{whitedmrg} to train Matrix Product States (MPS) \citep{Schollwockmps, ostlundmps} as a weight matrix in classifying \textit{MNIST} digits \citep{lecun2010mnist}. Subsequent work has also applied tensor networks in the classification of medical images \citep{selvan2020tensor} and regression \citep{reyes2020multiscale} but investigations into unsupervised and semi-supervised settings are lacking.

The literature on anomaly detection (AD) is vast and we will focus on reviewing previous work in the \textbf{one-class context} for \textbf{arbitrary data} (e.g. not restricted to images). Kernel-based methods, such as the One-Class SVM (OC-SVM) \citep{ocsvm}, learn a tight fit of inliers in an implicit high-dimensional feature space while the non-distance-based Isolation Forest \citep{isolationforest} directly distinguishes inliers and outliers based on partitions of feature values. Unfortunately, such classical AD algorithms presume the clustering of normal instances in some feature space and hence suffer from the curse of dimensionality, requiring substantial feature selection to operate on feature-rich, multivariate data \citep{bengio2012representation}.    

As such, hybrid methods were developed to first learn latent representations using Auto-Encoders (AE) \citep{xu2015aeocsvm, andrewsaeocsvm, Seebock_2019} and Deep Belief Networks (DBN) \citep{ERFANI2016121}, that were later fed to a OC-SVM. End-to-end deep learning models, without explicit AD objectives, have also been devised. Auto-Encoder AD models \citep{hawkinsae, sakuradaae, Chen2017OutlierDW} learn an encoding of inliers and subsequently use the reconstruction loss as a decision function. Other AE-variants, such as Deep Convolutional Auto-Encoders (DCAE) \citep{mascidcae, makhzanidcae}, have also been studied by \citep{Seebock_2019, Richter2017SafeVN}. Next, generative models learn a probability distribution for inliers and subsequently identify anomalous instances as those with low probabilities or those which are difficult to find in their latent spaces (in the case of latent variable models). Generative Adversarial Networks (GANs) \citep{goodfellow2014generative} have been popular in the latter category, with the advent of AnoGAN \citep{schlegl2017unsupervised}, a more efficient variant \citep{zenati2018efficient} based on BiGANs \citep{donahue2016adversarial}, GANomaly \citep{akcay2018ganomaly} and ADGAN \citep{adgan}. 

Deep learning models with objectives that resemble shallow kernel-based AD algorithms have also been explored. Such models train neural networks as explicit feature maps while concurrently finding the tightest decision boundary around the transformed training instances in the output space. Deep SVDD (DSVDD) \citep{pmlr-v80-ruff18a} seeks a minimal-volume hypersphere encapsulating inliers, motivated by the Support Vector Data Description (SVDD) \citep{taxsvdd}, while One-Class Neural Networks (OC-NN) \citep{chalapathy2018anomaly} searches for a maximum-margin hyperplane separating normal instances from the origin, in a fashion similar to OC-SVM. Contemporary attention has been directed towards self-supervised models, mostly for images \citep{geom, rotnet, hendrycks}, with the exception of the more recent GOAD \citep{goad} for general data. These models transform an input point into several altered instances according to a fixed class of rules and train a classifier that predicts a score for each altered instance belonging to its corresponding class of transformation. Outliers are then reflected as points with extreme scores, aggregated over all classes. In particular, GOAD unifies DSVDD and the self-supervised GEOM model \citep{geom} by defining the anomaly score of each transformed instance as its distance from its class' hypersphere center.

\section{Model Description}
\subsection{Overview}
In this section, we introduce our model that we call Tensor Network Anomaly Detector (TNAD). TNAD, as an end-to-end network with an explicit AD objective, falls into the second last category of models with one crucial caveat. Its notion of tightness does not rely on the volume of the decision boundary, which is an inadequate measure in the one-class setting. To illustrate this point, DSVDD and GOAD may find tiny hyperspheres during training but still have a loose fit around inliers, as they may map all possible inputs into the hyperspheres---a problem acknowledged by the original authors of DSVDD as ``hypersphere collapse'' \citep{taxsvdd}. In the scenario where outliers are available, one can indeed judge the tightness of a fit by the separation of inliers and outliers with respect to a decision boundary but in the one-class setting where no such points of reference are available, the tightness of a model's fit on training instances must be gauged relative to its other predictions. As such, we design TNAD to incorporate a canonical measure of its overall tendency to predict normality.



\begin{figure}[H]
\centering
\includegraphics[width=0.65\textwidth]{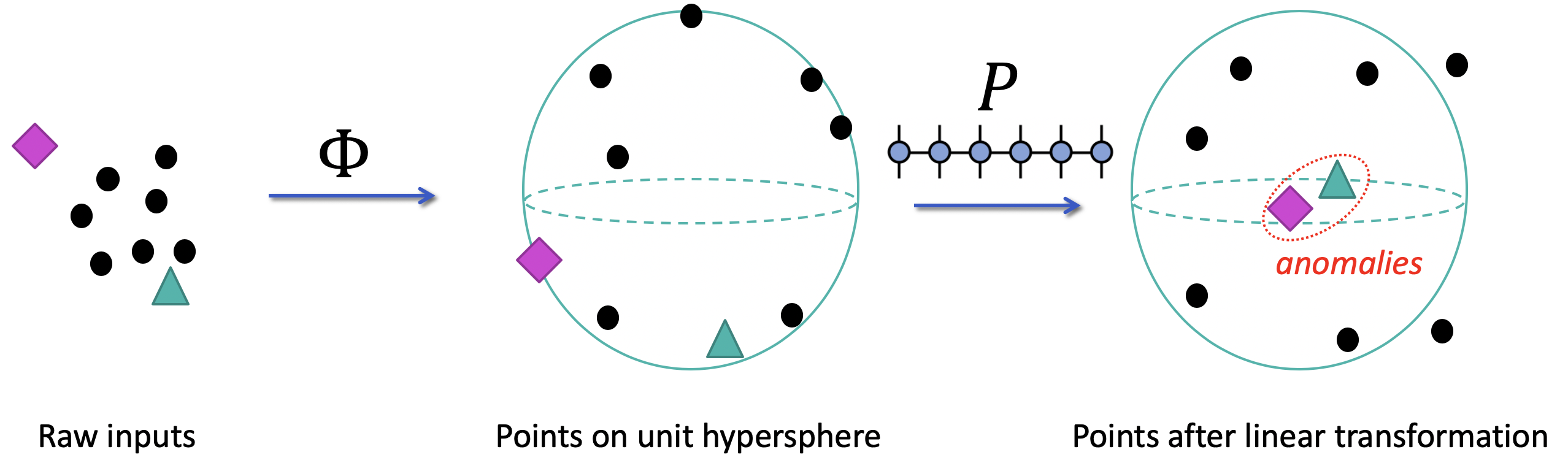}
\caption{Schematic of Tensor Network Anomaly Detector (TNAD)}
\label{TNAD}
\end{figure}

A schematic of TNAD is depicted in Figure \ref{TNAD}. A fixed feature map $\Phi$ is applied to map inputs onto the surface of a unit hypersphere in a vector space $V$ with dimension in the number of original features $N$. The training instances are sparse in this high-dimensional space $V$ and thus enables the learnt component of our model to be expressive, despite being a simple linear transformation $P: V \to W$. Upon action by $P$, normal instances will be mapped close to the surface of a hypersphere in $W$ of an arbitrarily chosen radius (set to be $\sqrt{e}$ in our experiments) while anomalous instances can be identified as those close to the origin. The decision function of the model with respect to an input $\bm{x}$, where a larger value indicates normality, is then
\begin{equation}
\mathcal{D}(\bm{x}) = \norm{P\Phi(\bm{x})}_2^2
\end{equation}
To accommodate the possible predominance of outliers, we allow $\dim W$ to have a smaller exponential scaling with $N$ so that $\dim W \ll \dim V$ for $P$ to have a large null-space. $P$ can thus be understood informally as a ``projection'' that annihilates the subspace spanned by outliers. To parameterize $P$ which has dimensions exponential in $N$, we leverage the Matrix Product Operator (MPO) tensor network, which is both memory and computationally-efficient. Finally, to obtain a tight fit around inliers, we penalize the Frobenius norm of $P$ during training.
\begin{equation}
\norm{P}_F^2 = tr\left(P^TP\right) = \sum_{i, j} |P_{ij}|^2
\end{equation}where $P_{ij}$ are the matrix elements of $P$ with respect to some basis. Since $\norm{P}_F^2$ is the sum of squared singular values of $P$, it captures the total extent to which the model is likely to deem an instance as normal. Ultimately, such a spectral property reflects the overall behavior of the model, rather than its restricted behavior on the training set.

\subsection{Matrix Product Operator Model}
In this section, the details of TNAD is expounded in tensor network notation---for which the reader is recommended to consult a brief introduction in our supplementary material and more comprehensive reviews in \citep{nutshelltn, practicaltn}. The input space $\mathcal{I}$ is assumed to be $[0, 1]^N$ for (flattened) grey-scale images and $\mathbb{R}^N$ for tabular data, where $N$ is the number of features. Given a predetermined map $\phi: \mathbb{R} \to \mathbb{R}^p$ where $p \in \mathbb{N}$ is a parameter known as the \textbf{physical dimension}, an input $\bm{x} = (x_1, ..., x_N) \in \mathcal{I}$ is first passed through a feature map $\Phi: \mathcal{I} \to V = \otimes_{j=1}^N \mathbb{R}^p$ defined by
\begin{equation}
\Phi(\bm{x}) = \phi(x_1) \otimes \phi(x_2) \otimes ... \otimes \phi(x_N)  
\end{equation}
Recall $\otimes_{j=1}^N \mathbb{R}^p$ is a $p^N$-dimensional and thus very large vector space. In tensor network notation, 
\begin{figure}[h]
\centering
\includegraphics[width=0.44\textwidth]{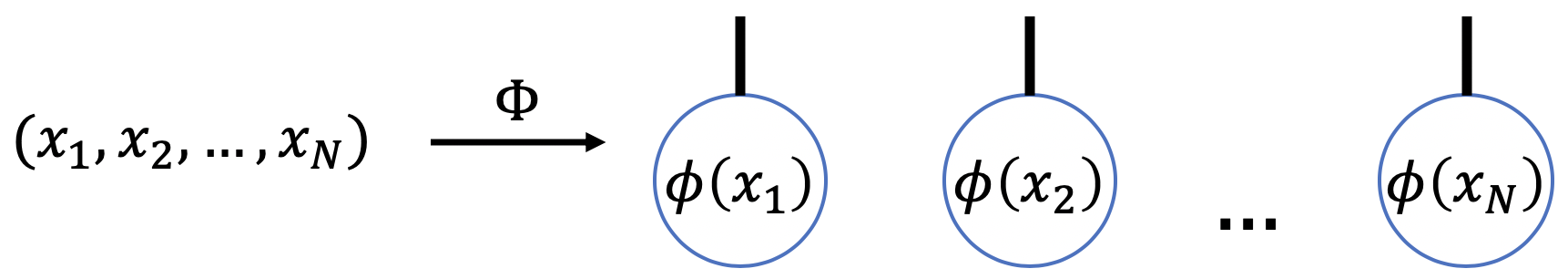}
\caption{TNAD Embedding layer.}
\label{embedding}
\end{figure}

The map $\phi$ is chosen to satisfy $\norm{\phi(y)}_2^2 = 1$ for all $y \in \mathbb{R}$ such that $\norm{\Phi(\bm{x})}_2^2 = \prod_{i=1}^N \norm{\phi(x_i)}_2^2 = 1$ for all $\bm{x} \in \mathcal{I}$, implying that $\Phi$ maps all points to the unit hypersphere in $V$. Two forms of $\phi$ were explored. The $2k$-dimensional ``trigonometric'' embedding $\phi_{trig}: \mathbb{R} \to \mathbb{R}^{2k}$ is defined as
\begin{equation}
\phi_{trig}(x) = \frac{1}{\sqrt{k}}\left(
\cos\left(\frac{\pi}{2}x\right), \sin\left(\frac{\pi}{2}x\right), ..., \cos\left(\frac{\pi}{2^k}x\right), \sin\left(\frac{\pi}{2^k}x\right)\right) 
\end{equation}
Our grey-scale image experiments were conducted with $\phi_{trig}$ with $p = 2k= 2$, which possesses the following natural interpretation. Since $\phi_{trig}(0), \phi_{trig}(1)$ are the two standard basis vectors $\bm{e}_1, \bm{e}_2$ of $\mathbb{R}^2 = \mathbb{R}^p$, the set of binary-valued images $\mathcal{B} = \{\bm{x} \in \mathcal{I}: \, \, x_i \in \{0, 1\} \, \, \forall 1 \leq i \leq N \}$ is mapped to the standard basis of $V$. Intuitively, the values $0$ and $1$ correspond to extreme cases in a feature (which reflects the pixel brightness in this case) so $\phi(0), \phi(1)$ are devised to be orthogonal for maximal separation. Now, for any $\bm{x}, \bm{y} \in \mathcal{I}$, since the inner product $\langle \Phi(\bm{x}), \Phi(\bm{y}) \rangle$ satisfies $\langle \Phi(\bm{x}), \Phi(\bm{y}) \rangle = \prod_{1 \leq i \leq N} \langle \phi(x_i), \phi(y_i) \rangle$, $\Phi$ is highly sensitive to each individual feature---flipping a single pixel value from $0$ to $1$ would lead to an orthogonal vector after $\Phi$. In essence, $\mathcal{B}$ then contains all extreme representatives of the input space $\mathcal{I}$, which can be seen to be images of highest contrast, and is mapped by $\Phi$ to the standard basis of $V$ for maximal separation. The squared F-norm of our subsequent linear transformation $P$ then obeys
\begin{equation}
\norm{P}_F^2 = \sum_{\bm{x} \in \mathcal{B}} \norm{P\Phi(\bm{x})}_2^2
\end{equation}
Recalling $\norm{P\Phi(\bm{x})}_2^2$ as the value of TNAD's decision function on an input $\bm{x}$, $\norm{P}_F^2$ is thus conferred the meaning of the total degree of normality predicted by the model on these extreme representatives---apt since images with the best contrast should be the most distinguishable. Unfortunately, such an interpretation of $\phi_{trig}$ does not extend to $k > 1$ so we also considered the $p$-dimensional ``fourier'' embedding $\phi_{four}: \mathbb{R} \to \mathbb{R}^p$ on tabular data, defined component-wise (indexing from $0$) as
\begin{equation}
\phi_{four}^j (x) = \frac{1}{p} \left| \sum_{k=0}^{p-1} e^{2\pi i k \left(\frac{p-1}{p}x - \frac{j}{p}\right)} \right|    
\end{equation}
This map has a period of $\frac{p}{p-1}$ and the following property. On $\left[0, \frac{p}{p-1}\right]$, the $i$-th value in $\{0, \frac{1}{p-1}, ..., \frac{p-2}{p-1}, 1 \}$ is mapped to the $i$-th standard basis vector of $\mathbb{R}^p$. Thus, $\{0, \frac{1}{p-1}, ..., \frac{p-2}{p-1}, 1 \}$ and its periodic-equivalents are deemed as extreme cases and a similar analysis follows as before. Ultimately, both versions of $\Phi$ segregate points close in the $L^2$-norm of the input space $\mathcal{I}$ by mapping inputs into the exponentially-large space $V$, buttressing the subsequent linear transformation $P$.

After the feature map, we learn a tensor $P_{i_1...i_q}^{j_1...j_N}: V  \to W = \otimes_{j=1}^q \mathbb{R}^p $ where $q = \floor{\frac{N-1}{S}}+1$ for some parameter $S \in \mathbb{N}$ referred to as the \textbf{spacing}. Our parameterization of $P$ in terms of rank-3 and 4 tensors is the below variant of the Matrix Product Operator (MPO) tensor network.  
\begin{figure}[h]
\centering
\includegraphics[width=0.55\textwidth]{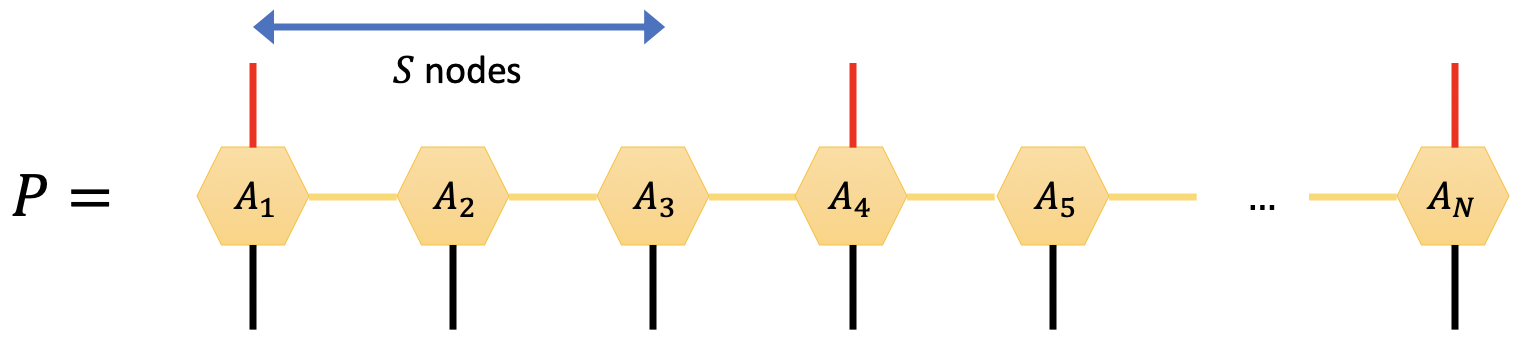}
\caption{Matrix product operator parameterization for $P$.}
\label{mpo}
\end{figure}

The modified MPO only has an outgoing red leg every $S$ nodes, beginning from the first. The red legs again have dimension $p$ while the gold legs have dimension $b$, which is another parameter known as the $\textbf{bond dimension}$. Intuitively, the gold legs are responsible for capturing correlations between features, for which a larger value of $b$ is desirable. In explicit tensor indices,
\begin{equation}
P_{i_1 ... i_q}^{j_1 ... j_N} = (A_1)_{i_1 k_1}^{j_1}(A_2)_{k_2}^{k_1 j_2}(A_3)_{ k_3}^{k_2j_3} ... (A_{S+1})_{i_2 k_{S+1}}^{k_{S}j_{S+1}}(A_{S+2})_{ k_{S+2}}^{k_{S+1}j_{S+2}}...
\end{equation}
where we have adopted Einstein's summation convention (see supplement) and $A_1, ..., A_N$ are the parameterizing low-rank tensors. TNAD's output, $\norm{P\Phi(\bm{x})}_2^2$, can then be computed as below.



\begin{figure}[H]
\centering
\includegraphics[width=0.6\textwidth]{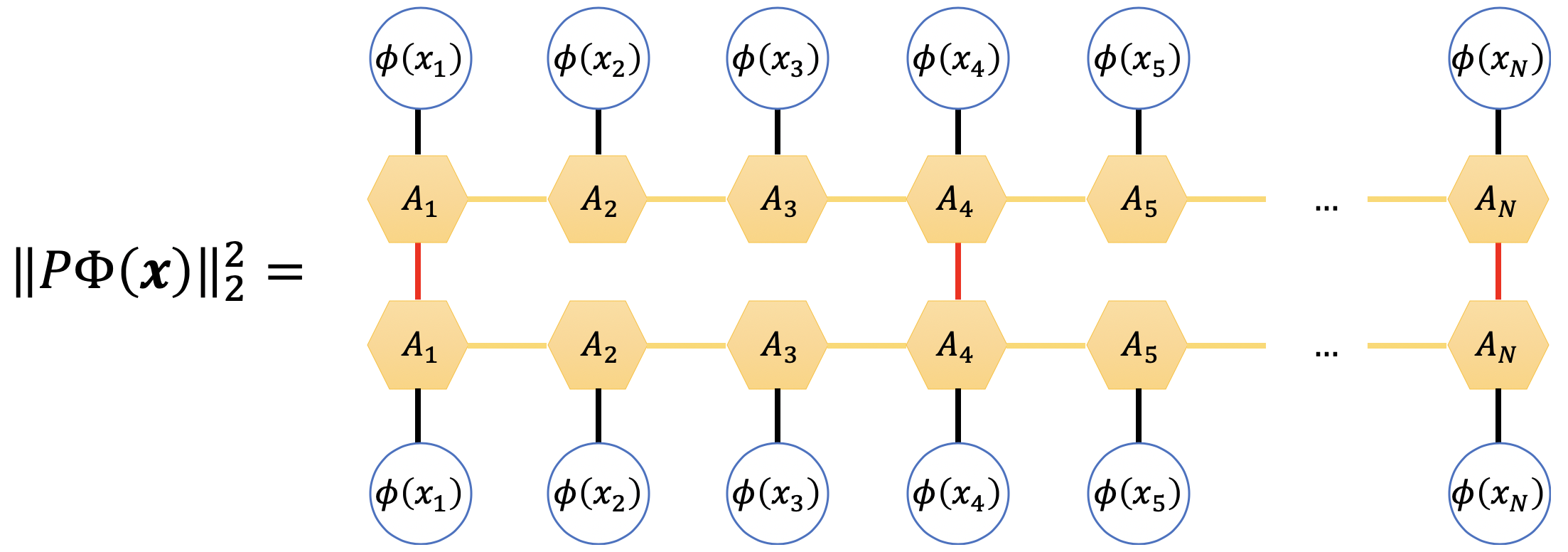}
\label{prediction}
\caption{Squared norm of transformed vector.}
\end{figure}

Finally, the following tensor network yields the squared F-norm of $P$ used as a training penalty.

\begin{figure}[H]
\centering
\includegraphics[width=0.55\textwidth]{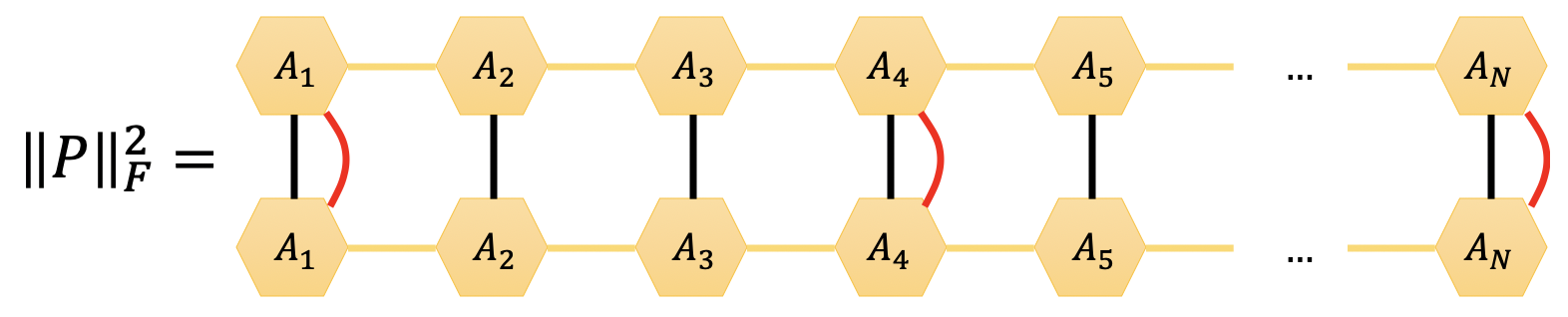}
\label{f-norm}
\caption{Squared F-norm of $P$.}
\end{figure}

Weaving the above together, our overall loss function over a batch of $B$ instances $\bm{x}_i$ is given by
\begin{equation}
\mathcal{L}_{batch} = \frac{1}{B} \sum_{i=1}^B \left(\log \norm{P\Phi(\bm{x}_i)}_2^2  - 1\right)^2+ \alpha \text{ReLU}(\log \norm{P}_F^2)
\end{equation}
where $\alpha$ is a hyperparameter that controls the trade-off between TNAD's fit around training points and its overall tendency to predict normality. In words, $P$ only sees normal instances during training which it tries to map to vectors on a hypersphere of radius $\sqrt{e}$, but it is simultaneously deterred from mapping other unseen instances to vectors of non-zero norm due to the $\norm{P}_F^2$ penalty. The $\log$'s are taken to stabilize the optimization by batch gradient descent since the value of a large tensor network can fluctuate by a few orders of magnitude with each descent step even with a tiny learning rate. Finally, the ReLU function is applied to the F-norm penalty to avoid the trivial solution of $P = 0$.

\subsection{Contraction Order and Complexity}
Now that our tensor network has been identified, the final ingredient is determining an efficient order for multiplying tensors---a process known as contraction---to compute $\norm{P\Phi(\bm{x})}_2^2$ and $\norm{P}_F^2$. Though different contraction schemes lead to the same result, they may have vastly different time complexities, for which the simplest example is the quantity $\norm{Av}_2^2 = v^T(A^TA)v = (Av)^T(Av)$ for some matrix $A$ and vector $v$---the first bracketing involves an expensive matrix product while the second bypasses it. The time-complexity of a contraction between two nodes can be read off a tensor network diagram as the product of the dimensions of all legs connected to the two nodes, without double-counting. Though searching for the optimal contraction order of a general network is NP-hard \citep{contractnphard}, the MPO has been extensively studied and an efficient contraction order that scales linearly with $N$ is known---despite being a linear transformation between spaces with dimensions exponential in $N$. The initials steps in computing $\norm{P\Phi(\bm{x})}_2^2$ are vertical contractions of the black legs followed by right-to-left horizontal contractions along segments between consecutive red legs.

\begin{figure}[h]
\centering
\includegraphics[width=1.0\textwidth]{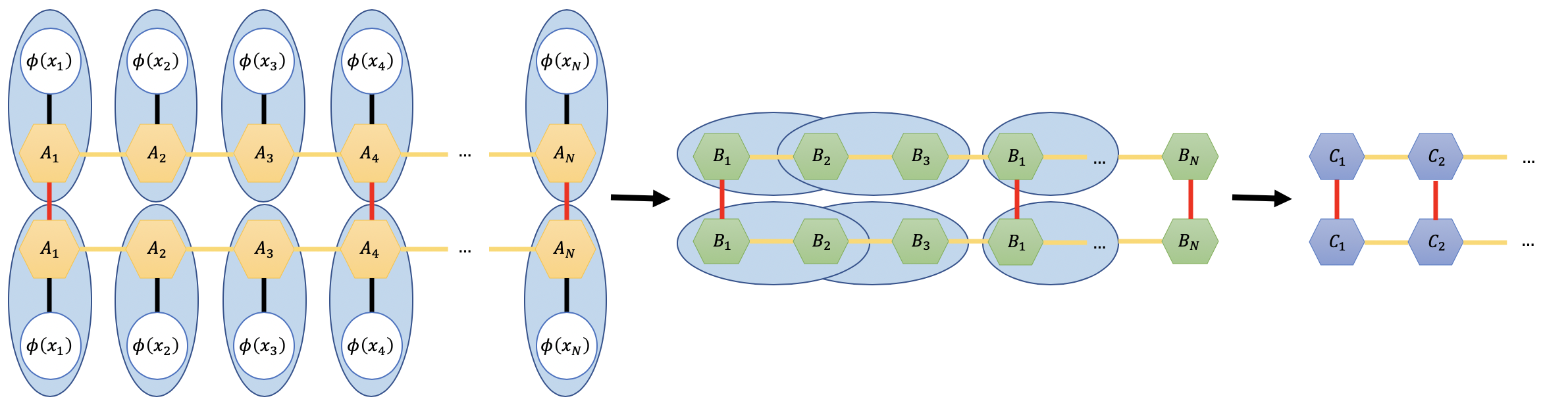}
\label{initialstep}
\caption{Initial step in computing $\norm{P\Phi(\bm{x})}_2^2$}
\end{figure}

\begin{figure}[h]
\centering
\includegraphics[width=0.55\textwidth]{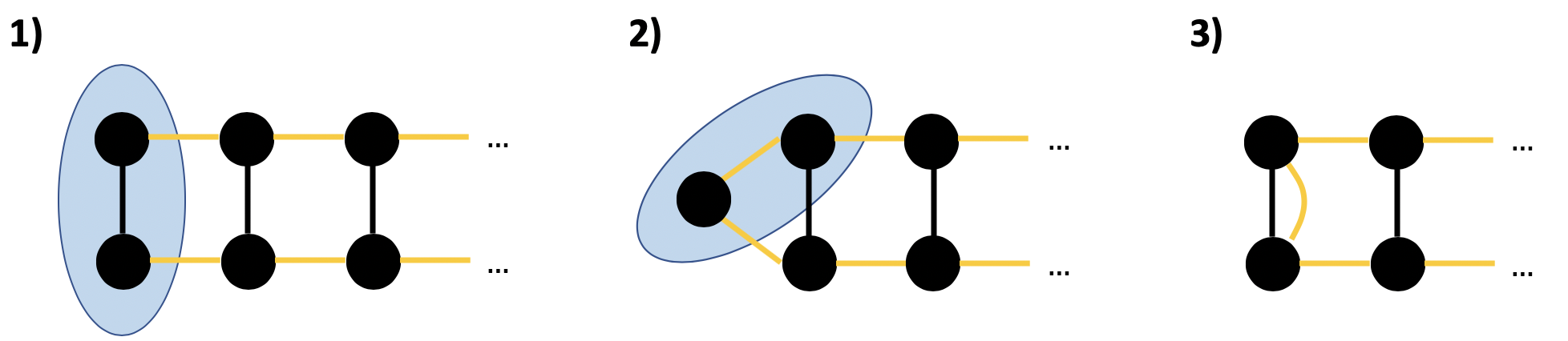}
\caption{Zig-zag contraction that is repeated till completion.}
\label{mpscontract}
\end{figure}

In practice, only the bottom half of the network is contracted before it is duplicated and attached to itself. Notably, this process can also easily be parallelized. At this juncture, observe that both $\norm{P}_F^2$ and the resulting network for $\norm{P\Phi(\bm{x})}_2^2$ are of the form in Figure \ref{mpscontract}, which can be computed efficiently by repeated zig-zag contractions. The overall time complexities of computing $\norm{P\Phi(\bm{x})}_2^2$ and $\norm{P}_F^2$ are $O\left(Nb^2 (b+p)  \left(\frac{p}{S} +1 \right)\right)$ and $O\left(N b^3 p \left(\frac{p}{S} + 1 \right)\right)$, where only the former is needed during prediction. Meanwhile, the overall space complexity of TNAD is $O\left(N b^2p \left(\frac{p}{S} + 1 \right)\right) $.

\section{Experiments}
The effectiveness of TNAD as a general one-class anomaly detector is verified on both image and tabular datasets. The Area Under the Receiver Operating Characteristic curve (AUROC) is adopted as a threshold-agnostic metric for all experiments. TNAD was implemented with the TensorNetwork library \citep{roberts2019tensornetwork} with the JAX backend and trained with the ADAM optimizer in its default settings.

\textit{General Baselines: }The general anomaly detection baselines evaluated are: One-Class SVM (OC-SVM) \citep{ocsvm}, Isolation Forest (IF) \citep{isolationforest}, and GOAD \citep{goad}. OC-SVM and IF are traditional anomaly detection algorithms known to perform well on general data while GOAD is a recent, state-of-the-art self-supervised algorithm with different transformation schemes for image and tabular data. OC-SVM and IF were taken off-shelf from the Scikit-Learn toolkit \citep{scikit-learn} while GOAD experiments were run with the official code of \citep{goad}. For all OC-SVM experiments, the RBF kernel was used and a grid sweep was conducted for the kernel coefficient $\gamma \in \{2^{-10}, ..., 2^{-1} \}$ and the margin parameter $\nu \in \{0.01, 0.1 \}$ according to the test set performance in hindsight---providing OC-SVM a supervised advantage. For all IF experiments, the number of trees and the sub-sampling size were set to $100$ and $256$ respectively, as recommended by the original paper. GOAD parameters are reported in the specific subsections.

\subsection{Image Experiments}
\textit{Datasets: }Our image experiments were conducted on the \textit{MNIST} \citep{lecun2010mnist} and \textit{Fashion-MNIST} \citep{xiao2017fashionmnist} datasets, each comprising $60000$ training and $10000$ test examples of $28 \times 28$ grey-scale images belonging to ten classes. In each set-up, one particular class was deemed as the inliers and all original training instances corresponding to that  class were retrieved to form the new training set, containing roughly $6000$ examples. The trained models were then evaluated on the untouched test set. 

\textit{Additional Image Baselines: }To illustrate the strengths of our approach, we include further comparisons to Deep SVDD (DSVDD) \citep{taxsvdd} and ADGAN \citep{adgan}, which entail convolutional networks. DSVDD experiments were performed with the original code while ADGAN results are reported from \citep{adgan, geom}.

\textit{Preprocessing: }For all models besides DSVDD, the pixel values of the grey-scale images were divided by $255$ to obtain a float in the range $[0, 1]$. Due to the computational complexity of TNAD, a $(2, 2)$-max pool operation with stride $(2, 2)$ was also performed to reduce the size of the images to $14 \times 14$ only for our model. In the cases of TNAD, OC-SVM and IF, the images were flattened before they were passed to these models---which thus do not exploit the inherent locality of the images, as contrasted with the convolutional architectures employed by all other models. For GOAD, the images were zero-padded to size $32 \times 32$ to be compatible with the official implementation designed for \textit{CIFAR-10}. Finally, for DSVDD, the images were preprocessed with global contrast normalization in the $L^1$-norm and subsequent min-max scaling to the interval $[0, 1]$, following the original paper.

\textit{Baseline Parameters: }The convolutional architectures and hyper-parameters of all deep baselines (GOAD, DSVDD, ADGAN) follow their original work. GOAD was run with the margin parameter $s=1$ and the geometric transformations of GEOM \citep{geom} involving flips, translations and rotations. DSVDD was run with $\nu = 0.1$ and a two-phased training scheme as described in the original paper. 

\textit{TNAD Parameters: }TNAD was run with physical dimension $p=2$, the $p$-dimensional embedding $\phi_{trig}$, bond-dimension $b=5$, spacing $S=8$, sites $N = 14 \times 14 = 196$ and margin strength $\alpha = 0.4$. All tensors were initialized via a normal distribution with standard deviation $0.5$. As an aside, TNAD is sensitive to initialization for large $N$ since it successively multiplies many tensors, causing the final result to vanish or explode if each tensor is too small or big---we found a standard deviation of $0.5$ to be suitable for $N=196$ and the final performance of TNAD to not vary significantly with the standard deviation, once it was initialized in a reasonable regime. As a further precaution, TNAD was first trained for $20$ ``cold-start'' epochs with learning rate $2 \times 10^{-5}$ to circumvent unfortunate initializations and a subsequent $280$ epochs with initial learning rate $2 \times 10^{-3}$ decaying exponentially at rate $0.01$, for a total of $300$ epochs. A small batch size $B=32$ was used due to space constraints. Finally, since only the $\log$'s of tensor network quantities are desired, we employ the trick of rescaling tensors by their element of largest magnitude during contractions and subsequently adding back the $\log$ of the rescaling to stabilize computations.

\textit{Results and Discussion: }Our experimental results on image datasets are presented in Table \ref{imageresults}. The mean AUROC, along with the standard error, across ten successful trials are reported for each model and each class chosen as the inliers. Occasionally, GOAD experienced ``hypersphere collapse'' while TNAD encountered numerical instabilities which led to \textit{nan} outputs---these trials were removed. Ultimately, TNAD produces consistently strong results and notably emerges second out of all evaluated models on \textit{MNIST}, surpassing all convolutional architectures besides GOAD despite not exploiting the innate structure of images. Furthermore, TNAD shows the lowest variation in performance other than the deterministic OC-SVM, possibly attributable to its linearity. OC-SVM exhibits a comparably strong performance though it was admittedly optimized in hindsight. Attaining the highest AUROC on most classes, GOAD undeniably triumphs all other evaluated models on images. However, GOAD's performance dip on \textit{MNIST} digits $\{0, 1, 8 \}$, which are largely unaffected by the horizontal flip and $180^{\circ}$ rotation used in its transformations, suggests that its success relies on identifying transformations that leverage the underlying structure of the data. Indeed, its authors \citep{goad} acknowledge that the random affine transformations used in GOAD's tabular experiments degraded its performance on image datasets. As such, TNAD's performance is especially encouraging, considering its ignorance of the inputs being images.

\begin{table}[!ht]
\centering 
\caption{Mean AUROC scores (in $\%$) and standard errors on \textit{MNIST} and \textit{Fashion MNIST}} \medskip
\begin{tabular}{cccccccc}
  \toprule
  Dataset & $c$ & SVM & IF & GOAD & DSVDD & ADGAN & TNAD \\
  \midrule
  \multirowcell{10}{\textit{MNIST}}& 0  & $\bm{99.5}$ & $96.4 \pm 0.6$ & $98.4 \pm 0.4$ & $98.2 \pm 0.6$ & $\bm{99.5}$  &  $99.2 \pm 0.0$ \\
  & 1  & $\bm{99.9}$  & $99.4 \pm 0.1$ & $96.5 \pm 0.9$ & $99.6 \pm 0.1$ & $\bm{99.9}$ & $99.8 \pm 0.0$\\
  & 2  & $92.6$ & $75.1 \pm 1.8$ & $\bm{99.6 \pm 0.0}$ & $90.3 \pm 2.5$ & $\bm{93.6}$ &  $92.7 \pm 0.3$\\
  & 3  & $93.8$ & $83.0 \pm 1.2$ & $\bm{98.6 \pm 0.2}$ & $90.1 \pm 2.3$ & $92.1$ &  $\bm{96.0 \pm 0.3}$\\
  & 4  & $\bm{97.1}$ & $87.0 \pm 0.9$ & $\bm{99.2 \pm 0.2}$ & $94.5 \pm 1.1$ & $93.6$  & $94.9 \pm 0.3$\\
  & 5  & $\bm{95.5}$ & $74.8 \pm 0.8$  & $\bm{99.5 \pm 0.1}$ & $87.1 \pm 1.4$ & $94.4$ &  $95.1 \pm 0.3$\\
  & 6  & $98.8$ & $86.9 \pm 0.9$ & $\bm{99.9 \pm 0.0}$ & $98.8 \pm 0.3$ & $96.7$ & $\bm{98.9 \pm 0.0}$ \\
  & 7  & $96.6$ & $91.2 \pm 0.7$ & $\bm{98.2 \pm 0.5}$ & $94.9 \pm 0.6$ & $96.8$  & $\bm{97.1 \pm 0.3}$\\
  & 8  & $90.8$ & $73.7 \pm 1.1$ & $\bm{96.9 \pm 0.4}$ & $93.3 \pm 1.1$ & $85.4$ & $\bm{94.9 \pm  0.3}$ \\
  & 9  & $96.3$ & $88.1 \pm 0.6$ & $\bm{99.0 \pm 0.2}$ & $96.3 \pm 0.9$ & $95.7$   & $\bm{97.2 \pm 0.1}$  \\
  \cdashline{2-8} 
  & avg  & $96.1$ & $84.6$ & $\bm{98.6}$ & $94.3$ & $94.7$ & $\bm{96.6}$\\
  \midrule
  \multirowcell{10}{\textit{Fashion-} \\\textit{MNIST}}& 0 & $91.9$ & $91.0 \pm 0.2$ & $\bm{93.4 \pm 0.6}$ & $90.1 \pm 0.8$ & $89.9$  & $\bm{92.5 \pm 0.2}$ \\
  & 1  & $\bm{99.0}$ & $97.6 \pm 0.1$ & $98.6 \pm 0.2$ & $\bm{98.7 \pm 0.1}$ & $81.9$   &  $97.5 \pm 0.1$ \\
  & 2  & $89.4$ & $87.1 \pm 0.4$ & $\bm{90.4 \pm 0.6}$ & $88.1 \pm 0.7$ & $87.6$  & $\bm{90.6 \pm 0.1}$ \\
  & 3  & $\bm{94.2}$ & $93.2 \pm 0.3$ & $91.0 \pm 1.5$ & $\bm{93.4 \pm 1.0}$ & $91.2$ & $91.8 \pm 0.2$ \\
  & 4  & $90.6$ & $90.2 \pm 0.5$ & $\bm{91.4 \pm 0.4}$ & $\bm{91.8 \pm 0.5}$ & $86.5$ & $90.5 \pm 0.1$\\
  & 5  & $91.8$ & $\bm{92.8 \pm 0.2}$ & $\bm{94.7 \pm 0.7}$ & $89.1 \pm 0.7$ & $89.6$ & $87.5 \pm 0.3$ \\
  & 6  & $\bm{83.5}$ & $79.5 \pm 0.6$ & $\bm{83.2 \pm 0.6}$ & $80.3 \pm 0.8$ & $74.3$ & $82.7 \pm 0.1$ \\
  & 7  & $\bm{98.8}$ & $98.3 \pm 0.1$ & $98.3 \pm 0.5$ & $98.4 \pm 0.2$ & $97.2$  & $\bm{98.9 \pm 0.0}$ \\
  & 8  & $89.9$ & $88.5 \pm 0.9$ & $\bm{98.8 \pm 0.2}$ & $\bm{92.9 \pm 1.3}$ & $89.0$  & $92.0 \pm 0.4$ \\
  & 9  & $98.2$ & $97.6 \pm 0.3$ & $\bm{99.3 \pm 0.2}$ &  $\bm{99.0 \pm 0.1}$ & $97.1$  & $97.8 \pm 0.2$ \\
  \cdashline{2-8} 
  & avg  & $\bm{92.7}$ & $91.6$ & $\bm{93.9}$ & $92.2$ & $88.4$  & $92.2$ \\
  \bottomrule
\end{tabular}
\caption*{In each row, the $c$-th class is taken as the normal instance while all other classes are anomalies. The top two results in each experiment are highlighted in bold. OC-SVM, which is abbreviated as SVM above, did not show variations in performance once it has converged so no standard errors are reported. ADGAN's results were borrowed from \citep{adgan, geom} which did not include error bars.}
\label{imageresults}
\end{table}
\subsection{Tabular Experiments}
\textit{Datasets: }We evaluate TNAD and other general baselines on $5$ real-world ODDS \citep{odds} datasets derived from the UCI repository \citep{ucidata}: \textit{Wine}, \textit{Glass}, \textit{Thyroid}, \textit{Satellite}, \textit{Forest Cover}. These were selected to exhibit a variety of dataset sizes, features and anomalous proportions---detailed information regarding them is presented in Table \ref{tabinfo}. Following the procedure of \citep{goad}, all models were trained on half of the normal instances and evaluated on the other half plus the anomalies.

\textit{Preprocessing: }For all models and datasets, the data was normalized such that the training set had zero mean and unit variance in each feature.

\textit{Baseline Parameters: }GOAD employs random affine transformations with output dimension $r$ for self-supervision on tabular data and trains a fully-connected classifier with hidden size $h$ and leaky-ReLU activations. We adhere to the hyperparameter choices in the original paper, setting $r=64$, $h=32$ and $25$ training epochs for the large-scale dataset $\textit{Forest Cover}$ and $r=32$, $h=8$ and $1$ training epoch for all other smaller-scale datasets. Finally, we also train DAGMM \citep{dagmm} using its original \textit{Thyroid} architecture for epochs in $\{10000, 20000, 30000, 40000\}$ and report the best results.

\textit{TNAD Parameters: }The dimensions of the input and output spaces $V$ and $W$, which depend on the parameters $N$, $p$ and $S$, are crucial to TNAD. As the number of features $N$ varies across datasets, we choose $p$ and $S$ according to the following heuristics. We set $S = \floor{\frac{N}{25}} + 1$ and subsequently choose $p$ such that $10^4 \leq \dim W = p^{ \floor{\frac{N-1}{S}} + 1} \leq 10^{12}$, with a preference for smaller $p$ on smaller datasets. The first rule imposes an appropriate nullspace of $P$ while the second ensures that the dimension is large enough to exploit the prowess of tensor networks while concurrently small enough to prevent the model from overfitting. On the smallest datasets \textit{Wine} and \textit{Glass}, we set $\alpha = 0.3$ to avoid overfitting while the other datasets used $\alpha = 0.1$. The bond dimension was fixed at $b=5$ and the same two-phased training scheme is adopted as before for the small tensor networks in \textit{Wine} and \textit{Thyroid}. For the other larger models, lower learning rates of $5.0 \times 10^{-6}$ damped and $5.0 \times 10^{-4}$ undamped were used, with a $5.0 \times 10^{-4}$ decay rate, to stabilize training. A batch size of $32$ was ued for all datasets besides \textit{Forest Cover} which used $512$. The embedding $\phi_{trig}$ is used by default. A summary of TNAD parameters is provided in Table \ref{tabinfo}.

\textit{TNAD Parameters on Glass: }The default parameters did not work well on \textit{Glass} so we devised $\phi_{four}$. To fully leverage the orthogonality properties of $\phi_{four}$, we choose a large $p=16$ and in turn set $S=2$ to maintain $\dim W$ in the desired range. 


\textit{Results and Discussion: }Table \ref{tabresults} shows the mean AUROC, with standard errors, from our experiments. Due to the large variance caused by its stochastic nature, GOAD was run for $500$ trials on small-scale datasets and $100$ trials on $\textit{Forest Cover}$. All other models were run for $10$ trials. TNAD is the best performer across all datasets. Its drastic improvements over the best baseline on the smaller datasets \textit{Wine} and \textit{Glass} bear credence to the ability of the F-norm penalty in ensuring a tight fit around scarce inliers. GOAD's poorer performance on \textit{Satellite} and \textit{Forest Cover} supports the expectation that affine transformations may not suit general data. All-in-all, we believe TNAD to be the best AD model, when given no prior domain knowledge of the underlying data.

\begin{table}[!ht]
\centering 
\caption{Information about ODDS datasets, sorted by size, and TNAD parameters used.}
\medskip 
\begin{tabular}{ccccccccccc}
\toprule
\multirow{2}{*}{\raisebox{-\heavyrulewidth}{Dataset}} & \multirow{2}{*}{\raisebox{-\heavyrulewidth}{\# Train}} & \multicolumn{2}{c}{\# Test} & \multicolumn{6}{c}{TNAD Parameters}\\
\cmidrule(lr){3-4}
\cmidrule(lr){5-10}
& & Normal & Anomalous & $N$ & $p$ & $S$ & $\dim W$ & $\alpha$ & $lr$ \\
\midrule 
\textit{Wine} & 59 & 60 ($85.7\%$) & 10 ($14.3\%$) & 13 & 4 & 1 & 2.9e4 & 0.3 & 2e-3 \\
\textit{Glass}* & 102 & 103 ($92.0\%$) & 9 ($8.0\%$) & 9 & 16 & 2 & 1.0e6 & 0.3 & 5e-4 \\
\textit{Thyroid} & 1839 & 1840 ($95.2\%$) & 93 ($4.8\%$) & 6 & 6 & 1 &  4.7e4 & 0.1 & 2e-3 \\
\textit{Satellite} & 2199 & 2200 ($51.9\%$) & 2036 ($48.1\%$) & 36 & 4 & 2 & 6.9e10 & 0.1 & 5e-4 \\
\textit{Forest} & 141650 & 141651 ($98.1\%$) & 2747 ($1.9\%$) & 10 & 8 & 1 & 1.1e9 & 0.1 & 5e-4   \\
\bottomrule 
\end{tabular}
\caption*{* Only for the \textit{Glass} dataset, $\phi_{four}$ was used while $\phi_{trig}$ was adopted for the others.}
\label{tabinfo}
\end{table}

\begin{table}[!ht]
\centering 
\caption{Mean AUROC scores (in $\%$) and standard errors on ODDS datasets.}
\medskip
\begin{tabular}{cccccc}
\toprule
Dataset & OC-SVM & IF & GOAD & DAGMM & TNAD \\
\midrule 
  \textit{Wine} & $\bm{60.0}$ & $46.0 \pm 8.4$ & $48.2 \pm 24.7$ & $51.7 \pm 19.3$ & $\bm{97.3  \pm 4.5}$ \\
  \textit{Glass} & $\bm{62.0}$ & $57.2 \pm 1.6$ & $53.5 \pm 13.6$  & $52.5 \pm 12.9$ & $\bm{81.8 \pm 7.3}$ \\
  \textit{Thyroid}  & $98.8$ & $\bm{99.0 \pm 0.1}$ & $95.8 \pm 1.3$ & $88.8 \pm 6.8$ & $\bm{99.0 \pm 0.1}$\\
  \textit{Satellite} & $\bm{79.9}$ & $77.2 \pm 0.9$ & $60.6 \pm 5.3$  & $72.1 \pm 4.7$ & $\bm{81.3 \pm 0.5}$ \\
  \textit{Forest} & $\bm{97.7}$ & $71.7 \pm 2.6$ & $64.6 \pm 4.7$  & $60.9 \pm 8.9$ & $\bm{98.8 \pm 0.6}$ \\
\bottomrule 
\end{tabular}
\caption*{The top two results in each experiment, ran for $10$ trials, are highlighted in bold. OC-SVM did not show variations in performance once it has converged so no standard errors are reported.}
\label{tabresults}
\end{table}

\newpage

\section{Conclusion}
In this paper, we have introduced TNAD as an adept anomaly detection model for general data. To the best of our knowledge, it is the first instance of a tensor network model exceeding classical and deep methods. It should be remarked that specifically for images and videos, there exist more appropriate tensor networks, such as PEPS \citep{pepstn}, that capture higher-dimensional correlations. Ultimately, we hope to set a paradigm of using tensor networks as ``wide'' models and spur future work.

\section{Broader Impact}
Anomaly detection has many benefits in fraud prevention, network security, health screening and crime investigation---the last two have been explicitly demonstrated by our successes in the \textit{Thyroid} and \textit{Glass} datasets. That said, anomaly detection also has applications in areas such as surveillance monitoring, which tie in issues such as individual privacy. Furthermore, what is considered anomalous may be a reflection of our societal norms so caution must taken in ensuring that such technology do not propagate inherent biases in our society. 

\section{Acknowledgements}
Jinhui Wang would like to thank David Lin Kewei from Stanford University for helpful discussions regarding the ``fourier'' embedding. 

X, formerly known as Google[x], is part of the Alphabet family of companies, which includes Google, Verily, Waymo, and others (www.x.company).

\newpage

\bibliography{paper}

\end{document}